\pdfoutput=1

\documentclass[11pt]{article}

\usepackage{acl}

\usepackage{times}
\usepackage{booktabs}
\usepackage{latexsym}
\usepackage{paralist}
\usepackage{csquotes}
\usepackage{amsmath}
\usepackage{multirow}
\usepackage{tabularx}
\usepackage{amssymb}
\usepackage{subcaption}
\usepackage{booktabs}
\usepackage{balance}

\usepackage{setspace}

\usepackage[T1]{fontenc}
\usepackage[utf8]{inputenc}

\usepackage{microtype}

\usepackage{inconsolata}
\usepackage{svg}
\usepackage{comment}
\usepackage{xcolor}

\title{Using Synchronic Definitions and Semantic Relations to Classify Semantic Change Types}

\author{
Pierluigi Cassotti\textsuperscript{\includegraphics[height=1.0em]{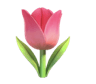}}, 
\textbf{Stefano De Pascale}\textsuperscript{\includegraphics[height=1.0em]{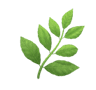}}, \textbf{and}
\textbf{Nina Tahmasebi}\textsuperscript{\includegraphics[height=1.0em]{got.png}}
 \\
  \textsuperscript{\includegraphics[height=0.85em]{got.png}}University of Gothenburg,
  \textsuperscript{\includegraphics[height=0.85em]{luv.png}}VUB/FWO/KU Leuven\\
  {\tt \{pierluigi.cassotti,nina.tahmasebi\}@gu.se}\\
{\tt stefano.de.pascale@vub.be }}

\begin{document}
\maketitle
\begin{abstract}

There is abundant evidence of the fact that the way words change their meaning
can be classified in different \textit{types} of change, highlighting the relationship between the old and new meanings (among which generalization, specialization and co-hyponymy transfer).
In this paper, we present a way of detecting these types of change by constructing a model that leverages information both from synchronic lexical relations and definitions of word meanings. Specifically, we use synset definitions and hierarchy information from WordNet and test it on a digitized version of Blank's (1997) dataset of semantic change types. Finally, we show how the sense relationships can improve models for both approximation of human judgments of semantic relatedness as well as binary Lexical Semantic Change Detection.

\end{abstract}


\section{Introduction}

At any point in time, a word can have several meanings. Often, these meanings share a certain degree of relationship with one another and the word is then said to display \textit{polysemy}. 
Diachronically, polysemy is a result of \textit{semantic change}, the process in which a word change its meaning/s. When new meanings are derived, the relationship to existing meanings is determined by the type of change that takes place. Examples include metaphorical extensions (\textit{the \textbf{arm }of the sea}) and specialization of usage (\textit{sand} that used to mean the shore as well as the grain of sand, and now only the latter).  
Several taxonomies of semantic change have been proposed, from \citet{reisig1839professor} 
to 
\citet{blank2012prinzipien}, that describe both the type of change as well as the causes of it. 

However, the computational community, which has spent the past 15 years developing computational models for automatically detecting semantic change in corpora, has disregarded those taxonomies and processes proposed in the past \cite{tahmasebi2021survey}. Sparked by the invention of neural word embedding techniques, the focus has been on quantifying the \textit{degree of change}, or determining \textit{when} change has taken place, without the use of sense-inventories, dictionaries or knowledge-bases. Currently, the standard recipe for lexical semantic change detection (LSC) is form-based \cite{montanelli2023survey}:  each usage of a word is represented by its contextualized embedding, and the embeddings of different time periods  are compared using cosine similarity. The comparison is done either by averaging pairwise similarities, or first averaging the embeddings of each time periods and then comparing the averages. When sense-based models are employed, (a) the embeddings are  clustered to derive sense representations and then clusters are compared over time, and optionally (b) the clusters are labeled to ease interpretation.  

So far, no computational work has been done to \textit{detect the qualitative type of change} that words experience. Nor have these 
change types been utilised when building models for change detection. Our work therefore presents a first approach to addressing both of these tasks.  


We rely directly on generated \textit{definitions} and sidestep the standard recipe, i.e., to represent each usage of a word with its contextualized embedding. We consider a word's meaning, i.e., sense, as a \textit{latent construct} and the definition as one way to capture that latent construct.  The use of definitions to precisely describe the meaning of a word offers an ideal level of sense representation, that combines both precision and abstraction, as well as a way to summarize multiple usages. Thus, the use of definitions serves as a dimension reduction technique, projecting N$\times$M contextualized embeddings in two time periods to a significantly smaller set of n$\times$m definitions. By using a single computational step rather than two (steps (a) and (b) above), we reduce the complexity and  decrease the uncertainty. 

To classify semantic change types, we train a classifier on a \textit{synchronic} sense repository with definitions and semantic relations, namely WordNet. The rationale behind this choice is that each of the type of semantic change targeted in this paper corresponds to a synchronic relationship between senses. This distant training procedure via synchronic information is necessary as there are no available resources to date for change type classification that surpass a few hundred examples.

Next, we test this classifier on an unseen dataset of historical semantic changes. We therefore extend the dataset provided by \citet{blank2012prinzipien}, in which for a given word and a pair of senses there is an associated type of semantic change, with concise definitions for each sense. 
Finally, we incorporate change type classification in a state-of-the art model for detection of semantic change, and show an improvement in the model performance. 

\paragraph{Contributions: }In this paper, we show that
\begin{compactenum}
    \item definitions of word senses can be used to detect semantic change type;
    \item we can classify the type of semantic change by training on synchronic sense relationships using sense definitions; and that
    \item the use of type information can improve models for both graded Word-In-Context (WiC) as well as semantic change detection. 
\end{compactenum}

\section{Background and Related Work}
\subsection{Lexical Semantic Change Detection}
\label{lsc-sota}
Lexical Semantic Change Detection (LSC) is the task of automatically identifying words that change their meaning over time. Meaning changes that take place because whole senses are added or lost over time are considered as \textit{binary change}. Change can also be more subtle and relate to existing senses, for example, the decrease in frequency of a dominant sense. We consider the latter as \textit{graded change}. Words can experience both binary and graded change over (longer or shorter) periods of time, and both add some senses, loose some, and change others. 

Up until 2020, the majority of the methods that computationally modeled semantic change were tested only on the ability to say if a word had changed or not, conflating all kinds of change into one. Evaluation was ad-hoc and as a result, no comparability across different methods was possible. This changed when the SemEval-2020 Task 1 on Unsupervised Lexical Semantic Change Detection campaign \cite{schlechtweg2020semeval} was launched in 2020. As a part of the shared task,  benchmarks covering Swedish, English, German and Latin were released. Each benchmark covers a set of words, across a pair of corpora (or subcorpora) with a temporal gap between the two. For each word, up to a 100  usages are sampled from each time period. These usages are paired both within a corpus and across the corpora.  Each annotated pair receive a \textit{graded} Word-in-Context score (see Table \ref{en:scale}) by human annotators. The usages are considered \textit{nodes} and the pairwise annotations as \textit{edges} when a diachronic word usage graph (DWUG) \cite{schlechtweg-resource-2020-21} is created. The DWUG is clustered using correlation clustering and distribution of cluster members across time is used to attain the binary and graded change scores for each word. 
The information within a corpus facilitates the possibility to differentiate the synchronic senses (polysemy and homonymy), while comparing across corpora for the estimation of semantic change. 

\begin{table}[!ht]
\centering
\resizebox{0.8\columnwidth}{!}{%
\begin{tabular}{lcll}
\multirow{4}{*}{$\Bigg\uparrow$} 
& 4: & Identical & Identity\\
& 3: & Closely related & Context variance\\
& 2: & Distantly related & Polysemy\\
& 1: & Unrelated & Homonymy        
\end{tabular}}
\caption{The DURel relatedness scale \cite{schlechtweg2018diachronic} and the respective Continuum of semantic proximity proposed by \citet{blank2012prinzipien}.}%
\label{en:scale}
\end{table}

Methods for encoding word meaning in the LSC task range from distributional semantic models such as count-based or Word2Vec approaches \cite{hamilton2016cultural} and contextualized models \cite{laicher} to the most recent task-informed models such as XL-LEXEME \cite{cassotti-etal-2023-xl} and models based on definitions \cite{giulianelli-etal-2023-interpretable}. 
Task-informed models are supervised models leveraging information from different tasks for LSC prediction. XL-LEXEME and Deep Mistake \cite{arefyev_deepmistake_2021} are models trained 
on the WiC task, while Gloss Reader \cite{rachinskiy_zeroshot_2021} exploits the Word Sense Disambiguation task. Definition-based models, on the other hand, finetune LLMs for in-context definition generation. XL-LEXEME holds the state of the art in multiple languages \cite{periti2024chatgpt} and will thus be used here. 

\subsection{Sense Definition Modeling}
Highly related to our work is the task of Word Sense Disambiguation (WSD) that has long been of great interest to the community \cite{navigli_survey}. Similar to other NLP tasks, model architectures have evolved over time, including models based on knowledge-bases or the more common ones based purely on neural networks. With the outstanding performance on a majority of existing NLP tasks, generative Large Language Models (LLMs)  
have contributed to a 
paradigmatic shift also for WSD.
While previous models assign a pre-defined label to a given usage of word $w$, current models generate the label of the word meaning. 
In particular, such a label could be the definition of the word in the specific usage context.

In this work we refer to \textit{definition modeling} or \textit{definition generation} as the task of automatically generating dictionary-like definitions of a (latent) word sense given a certain a word usage. The task is carried out using several different approaches \cite{defmodelingsurvey} and datasets, such as WordNet \cite{princeton-wordnet}, Wiktionary \footnote{\url{https://www.wiktionary.org}}, Oxford English Dictionary (OED) \footnote{\url{https://www.oed.com}} and the Urban Dictionary \footnote{\url{https://www.urbandictionary.com}}. Noteworthy is the contribution of \citet{bevilacqua-etal-2020-generationary}, which is one of the first works using a large pre-trained model, followed by \citet{huang-etal-2021-definition} who introduce a way to control the specificity of the generated sense definitions, and finally, \citet{giulianelli-etal-2023-interpretable} who use similarities between word sense definitions to approximate human judgments on semantic proximity. We will use the latter model in our work.

\subsection{Classification in Historical Semantics}
The way in which semantic change manifests itself is of great interest to the linguistic community and over time various taxonomies of semantic change have been proposed \cite{reisig1839professor,paul2010prinzipien,darmesteter1893vie,breal1904essai,stern1931meaning,bloomfield1994language,ullmann,blank2012prinzipien}. These include a rich set of examples that have undergone semantic change. While there is a large overlap between the taxonomies (with generalization, specialization, metaphorical and metonymical changes as its recurring core) they are very much indebted to the then dominant theoretical frameworks (e.g., historical-philological semantics, structuralist semantics etc.) \cite{geeraerts2010}. All taxonomies have been created using traditional, qualitative approaches and while relying on collections of texts, the approach are not data-driven.

The lack of resources tailored to train models on these complex semantic changes has prevented computational work thus far. Semantic change often entails a complete loss of previous meanings, making most contemporary computational lexical resources insufficient, as they typically miss the historical meanings. Although datasets exist for phenomena like metaphors, metonymies, or analogies, they often concentrate on the novel use of word meanings and omit conventional examples \cite{maudslay-teufel-2022-metaphorical} thus limiting the models' ability to predict semantic change types. While there exists no resources for training models, in this paper, we release a first benchmark for testing models on their ability to detect semantic change types given definitions.

\section{LSC-CTD Benchmark}

One of the most comprehensive and recent classification of semantic change is Blank's taxonomy \cite{blank2012prinzipien}. In this work,  about 650 cases of semantic change are classified covering the vocabulary of the Romance languages and to some extent also German and English.  Blank tries to reinterpret traditional types of classification from a cognitive-linguistics point of view.  For the LSC Cause-Type-Definitions (LSC-CTD) Benchmark\footnote{The benchmark is available on Zenodo \url{https://zenodo.org/doi/10.5281/zenodo.11471317}} we digitize the cases reported by Blank and create a benchmark of definitions manually curated by an historical linguist assisted by ChatGPT\footnote{\url{https://chat.openai.com/}}. The format of the dataset, the dataset statistics and the annotation procedure are described further in Appendix \ref{app:LSC-def-bench}.

Blank's type classification include specialization, generalization, co-hyponymous transfer, auto-antonym, metaphor, antiphrasis, metonymy, auto-converse, ellipsis, folk-etymology, analogy, meaning dilution, meaning reinforcement, and doubtful cases. For our study, we use the following types of change for which we can find a counterpart in the synchronic lexical database.\footnote{Descriptions based on  \citet{Geeraerts_2020}.}

\begin{itemize}
  \item[ \textbf{\texttt{generalization}}:] the old meaning is a sub-case of the new meaning (or: the new meaning has at least one less definitional feature than the old meaning); other terms used to refer to this type of change are \textit{broadening} and \textit{widening}.  An example is the (non-attested) Vulgar Latin \textit{*adripare}, which originally meant 'to reach the shore (by vessel)' but in contemporary Romance languages has generalised to 'arrive (by any means of transport)' (e.g.: French \textit{arriver}, Italian \textit{arrivare}) 
  \item[\textbf{\texttt{specialization}}:] the new meaning is a sub-case of the old meaning (or: the new meaning has at least one extra definitional feature than the old meaning); other terms used to refer to this type of change are \textit{narrowing} and \textit{restriction}. An example is Latin \textit{necare}, which used to refer to 'to kill (by any means)' but whose French derivation \textit{noyer} is now restricted to 'to kill by drowning' 
  \item[\textbf{\texttt{co-hyponymous transfer}}:] the change that occurs due to naming confusion, that is, when a word is used for new referents that are (wrongly) thought to be similar if not identical to the old referents. For instance, in some Romance languages the word for 'rat' is also used (in a derived form) to refer to 'mouse' or vice versa (Piedmontese: \textit{rat} for 'rat' and \textit{rató} for 'mouse'; Spanish: \textit{ratón} for 'rat' and \textit{rata} for 'mouse')
  \item[\textbf{\texttt{auto-antonymy}}:] also called \textit{contronymy} or \textit{enantiosemy}, is when a word develops a new meaning that expresses a contrast, or is in opposition, to its old meaning. For instance, the Latin adjective \textit{sacer}  started out with the positive sense of 'sacred' and only later acquired the negative sense 'cursed', as evinced by the French \textit{sacré} \cite{traugott2017, karaman2008}
\end{itemize}

\section{Generating examples from WordNet}
\label{sec:wn_data}

%

To train a classifier, we use WordNet, a database that provides a large set of examples of lexical-semantic relations that are the synchronic counterpart of the semantic change types. WordNet is a multilingual synchronic lexical resource that can be queried to obtain semantic information for the covered vocabulary.  Each \textit{lemma} in WordNet is found within a \textit{synset}, which brings together cognitive synonyms. Each synset therefore represent a sense or concept shared between several lemmas. Synsets are organized hierarchically in a tree structure, where the arc between a node and its parent represents a hyponymy relationship, meaning the lower node represents a more specific concept than that represented by the parent. This relationship can also be viewed inversely, where each parent node is connected to a child node if and only if there is a hyperonymy relationship between them, that is, the parent expresses a more general concept than the child node.

WordNet associates each synset with a \textit{sense gloss} and a series of \textit{corpus examples}. The gloss functions as a concise definition which has been specifically conceived to uniquely express the meaning of the synset, and thus exemplify the latent construct of a sense. The examples, on the other hand, are mere uses of the words, which are often under-specified with respect to their meaning. For this reason, we focus on the sense gloss. 

The relations of hyperonymy, hyponymy, co-hyponymy and antonymy can all be considered the synchronic counterpart of the semantic change types introduced in Section 3. Below we illustrate how we map the taxonomical relationship between the sense glosses of synsets in WordNet to each considered change type. Change types in Blank's taxonomy that cannot be modelled through the WordNet hierarchy are disregarded for this work. For hierarchical relationships of hyponymy, hyperonymy, and co-hyponymy, we consider paths of maximum length 1, meaning pairs for which there is a direct relationship.
\parskip=0pt\relax

\paragraph{Generalization and specialization:} For every synset, we extract the list of associated hyperonymous synsets, and form pairs of a synset sense gloss with its hyperonym sense gloss. Viewed from the perspective of the child synset to the parent synset, we obtain instances of generalization. Similarly, by exploiting the inverse relationship (from parent to child synset), we obtain examples of specialization.

\paragraph{Co-hyponymy:} We define a co-hyponymy relationship as two separate synset nodes with a common synset parent, in which the two child synsets
represent two different and more specialized cases of the same general concept. By pairing the sense glosses of the child synsets
, we generate the same sense relationship that lies at the basis of the diachronic co-hyponymous transfer. 

\paragraph{Antonymy:} WordNet also encodes the relationship between synsets that express opposite concepts, often between adjectives, but also found between other word classes such as nouns (\textit{father} vs. \textit{mother}) and verbs (\textit{to buy} vs. \textit{to sell}). For each synset, we therefore extract the synset, or the lemma within the synset, labeled as antonym, and pair the sense glosses of these two synsets. 

\paragraph{Homonymy:} Finally, we randomly generate pairs of synsets with the same part-of-speech that do not lead to hyperonymous/hyponymous, co-hyponynic and antonymic pairings of sense glosses. 
The potential unrelatedness between the senses should be a good proxy for homonymy, 
defined as the occurrence of completely unrelated senses (for the same word form). While it is possible that related senses of the same word are paired in the random pairing, the risk is extremely small (i.e., on average 3-4 out of the 30-40,000 randomly sampled synsets can belong to the same word). Homonymy is not treated on the same par as the other types of semantic change by \citet{blank2012prinzipien}, even though its importance as a driver of language change is uncontroversial. For this reason it has been included in the experiments.


\begin{table}[h]
    \centering
    \small
\begin{tabular}{rl}\toprule
Semantic Change type & Synchronic relation \\
\midrule
Generalization & Hyperonymy\\
Specialization & Hyponymy\\
Co-hyponymous transfer & Co-hyponymy\\
Auto-antonymy & Antonymy\\
Unrelated & Homonymy\\\bottomrule
\end{tabular}
\caption{Mapping between Semantic change types to the synchronic relations found in WordNet.}\label{tab:mapping}
\end{table}
\paragraph{The WordNet testset:}

We divide the final dataset into three distinct datasets: training, validation, and testing, selecting  80\%, 10\%, and 10\% of the data respectively. The examples are split in such a way that the same pair of definitions for a specific class never appears simultaneously in two datasets (training/validation, validation/test, training/test). Furthermore, to avoid class imbalance, we limit the maximum number of examples per class to 30,000, 3,000, and 3,000 for training, validation, and testing respectively.
The statistic of the dataset are documented in Appendix \ref{app:wn_stats}.

\section{Model}
\label{sec:model}
To classify the relationship between senses, we utilize the data collected in Section \ref{sec:wn_data} and train a classification model\footnote{Our code is available on GitHub \url{https://github.com/ChangeIsKey/change-type-classification}. The classification model is available on the Hugging Face Model Hub \url{https://huggingface.co/ChangeIsKey/change-type-classifier}.} that takes as input the two definitions of the respective senses and outputs the corresponding class, i.e. \{\textit{homonymy, hyperonymy, hyponymy}, \textit{antonymy}\}. 

Specifically, we concatenate the textual definitions $\delta_1$ and $\delta_2$ together with special tokens into a single string $\delta = \text{<s>} \delta_1 + \delta_2 \text{</s>}$, which is then tokenized and encoded using \texttt{roberta-large} \cite{liuroberta2019}. From this, we obtain the vectors corresponding to the sub-tokens of the original string $v_1,..,v_K$. The vector $v_1$, corresponding to the initial token, is used for classification. We apply a linear transformation to reduce the vector's dimension to the label space, such that $v_1^* = Wv_1$, where $W \in \mathbb{R}^{C \times N}$ and $C$ is the number of classes. Finally, we use cross-entropy to optimize the model. 

To counteract class imbalance, examples in each batch are drawn proportionally from each class. We use AdamW for optimization with a weight decay of 0.1 and a learning rate of 1e-6. Early stopping is employed to determine when to halt training based on accuracy on the development set. Additionally, we assess the outcomes using two baselines. The results for these baselines are detailed in Appendix \ref{app:baseline}.

\section{Evaluation of the Type Classification}

We first evaluate the model on the test portion of the synchronic relations found in WordNet. Next, we test the model on a historical,  novel data, namely the LSC-CTD benchmark. 

\begin{figure}[h]
    \centering
    \includesvg[width=\columnwidth]{cfn_wn.svg}
    \caption{Confusion matrix for predictions of sense relationships on the WordNet testset.}
    \label{fig:confusion_matrix_wn}
\end{figure}
\subsection{Evaluation on WordNet}
We evaluate the classifier on the WordNet testset and investigate each class individually using the confusion matrix found in Figure \ref{fig:confusion_matrix_wn}.  The diagonal results show correctly classified predictions for each class. The normalized recall rate range from 0.83 for the co-hyponym class to 0.95 for the honomym class. That means, the model correctly identifies between 83\% -- 95\% of the true class members showing that the model performs well in distinguishing these classes.

For hyperonyms and hyponyms, the model shows a normalized recall rate of 0.91 and 0.90 respectively. The confusion mainly occurs with the converse relationship (hyponyms/hyperonyms) (0.048/0.057), which is logical given the close relationship between hyperonyms and hyponyms, as a hyperonym's definition is inherently included within its hyponym's definition. There are minor confusions with co-hyponyms (0.03) and a negligible amount with antonyms and homonyms. 

The model exhibits the lowest normalized recall rate of 0.83 for co-hyponyms. Despite being almost 10-points lower than that for hyperonyms and hyponyms, it still suggests a substantial capability to correctly classify co-hyponyms. The confusion is distributed across the other hierarchical relationships. Co-hyponyms get misclassified as hyperonyms (0.07), hyponyms (0.06), and to a very small proportion as homonyms (0.01). The slightly lower performance in this category could be due to the more complex nature of co-hyponym relationships, which require the model to understand not just a direct hierarchical link, but also parallel connections within the same level of hierarchy.

The negative class serves as a control or baseline class against which hierarchical or antonym relationships are compared. The model correctly identify 95\% of instances that do not exhibit a hierarchical relationship. 

The small fractions of misclassification with other classes (0.008 with hyponym, 0.009 with hyperonym, 0.026 with co-hyponym, and 0.004 with antonym) indicate that the model is very effective at distinguishing non-hierarchical and non-antonym relations from homonyms ones. This high performance is crucial as it shows the model's strong discriminatory power, effectively reducing the false positive rate where hierarchical or antonym relationships are incorrectly inferred. 

\subsection{Evaluation on the LSC-CTD Benchmark}
To answer how well the model trained on synchronic relations can classify diachronic relations, i.e., semantic change types, we test the model on the LSC-CTD Benchmark. We follow the mapping shown in Table \ref{tab:mapping} and consider a prediction correct within the mapped class. As an example, a pair of definitions classified as having exhibited specialization is correct if the model predicts hyponomy. Result are shown in Figure \ref{fig:confusion_matrix_blank}. 

\begin{figure}[h]
    \centering
    \includesvg[width=\columnwidth]{cfn.svg}
    \caption{Confusion matrix for predictions of sense relationships on the LSC-CTD Benchmark.}
    \label{fig:confusion_matrix_blank}
\end{figure}

The model shows a high normalized recall rate of 0.85 for Specialization, that is, 85\% of all the specialization instances are correctly predicted as hyponyms. The confusion occurs with co-hyponyms (0.15). This suggests that while the model is generally good at recognizing the specialization that occurs from a general to a more specific sense, it can sometimes fail to see the vertical relationships between these senses and consider them on the same horizontal level (as co-hyponyms). 

The model exhibits a normalized recall rate of 0.68 for generalization, i.e., moderate success in identifying instances of generalization. There is a notable amount of confusion with hyponyms (0.16), and co-hyponyms (0.11).
This asymmetry  between generalization and specialization classes could be due to the fact that going from more specific to more general definitions often implies the use of fewer and more abstract or polysemous terms, which are notoriously harder to represent. 

The model has a low normalized recall rate of 0.17 for co-hyponymous transfer, suggesting significant challenges in accurately classifying this type of semantic change. The confusion is well distributed across the other classes. with the largest misclassification being with hyponyms (0.46) and hyperonyms (0.29). There might be two reasons for this: co-hyponymous transfer fundamentally relies on confusion and referential ambiguity on the side of the speaker, so it is not surprising that specifically text-based LLMs cannot handle such confusion; in Blank's dataset it is considered a phenomenon of rather limited range, mainly observed in specific dialects (a third of the examples come from Réunion Creole).
The model seems to have a moderately good grasp of auto-antonymy correctly classifying  62\% of these instances. Here there is room for improvement, in particular by increasing the number of training examples on synchronic antonym relations.

\section{Evaluation on Semantic Change}
We will use our classification of semantic change type, introduced in Section \ref{sec:model}, to test if we can improve on state-of-the-art LSC detection. In this section, we use Flan-T5 XL FT presented by \citet{giulianelli-etal-2023-interpretable} to generate definitions, which is a Flan-T5 XL \cite{chung2022scaling} model finetuned on WordNet \cite{miller2013}, the Oxford dataset \cite{gadetsky-etal-2018-conditional} and the CoDWoE dataset \cite{DBLP:conf/semeval/MickusDCP22}.

\begin{figure}
    \centering
    \includegraphics[width=\columnwidth]{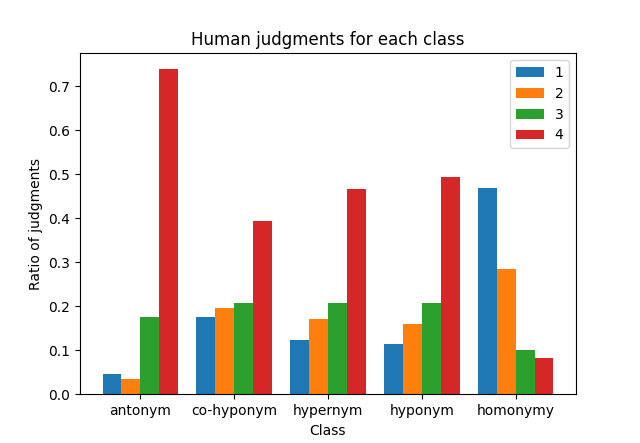}
    \caption{Ratio of human judgments for each class.}
    \label{fig:spearman-judgment}
\end{figure}

\subsection{Semantic Relatedness and Homonymy}
In this Section, we use the English portion of the SemEval-2020 Task 1 to study the correlation between human annotated relatedness scores and word sense relationships.

In the annotation of SemEval-2020 Task 1, (see further Section \ref{lsc-sota}), pairs of usages are sampled for each word. These usage pairs are either synchronic, i.e. belong to the same time period, or diachronic, i.e. belong to different time periods. For each pair of usages, the annotators have to assign a label on a scale from 1 to 4 (Table \ref{en:scale}): 1 (unrelated), 2 (distantly related), 3 (closely related), and 4 (identical). This scale draws inspiration from the work of \citet{blank2012prinzipien}, which introduced the concept of a continuum of semantic proximity. Specifically, \citet{schlechtweg2018diachronic} provides a direct mapping between the relatedness scale and the semantic proximity introduced by \citet{blank2012prinzipien}.

For each usage in the English portion of the SemEval-2020 Task 1, we generate a definition. We then predict word sense relationships  using the definitions for all the usage pairs for which human judgments are available. The word sense relationships include homonymy, co-hyponymy (co-hyponymous transfer) , hyperonymy (generalization), hyponymy (specialization) and antonymy (auto-antonymy). The bar graph in Figure \ref{fig:spearman-judgment} represents, for each predicted class, the distribution of the existing human judgments based on the semantic relatedness scale\footnote{It is important to note that the type classification model used in the study is not designed to identify identical meanings, which corresponds to a human judgment of 4.}.

The homonymy category has a high number of judgments in class 1 (Unrelated) (60\% of the overall unrelated pairs), which is consistent with the definition of homonyms: words that are spelled the same and sound the same but have different meanings. There are very few 3 or 4 judgments, which suggests that humans generally do not find homonyms to be semantically related.

Overall, the distribution of human judgments reflects an understanding of semantic relationships that aligns with the theoretical definitions of each relationship type. Closely related judgments dominate in categories where a semantic link is expected (co-hyponym, hyperonym, hyponym), while unrelated judgments  are more frequent for homonymy samples, where no semantic relationship or an opposite relationship is anticipated.

\subsection{Graded WiC Task}
The model's exceptional ability to identify unrelated pairs annotated with 1 clearly demonstrates its adequacy in modeling the homonymy class. We will use this ability to improve models for the Graded WiC task.

We set up the evaluation as a Graded WiC task, directly comparing the semantic proximity value provided by human annotators and the label computed by the model. To do this, we transform the model's computed labels into a two-value scale: 1 (Related) and 0 (Unrelated), grouping the labels hyperonym, hyponym, co-hyponym, and antonym under the Related class, while the homonymy class represents the Unrelated class. 

Furthermore, we investigate the combination of XL-LEXEME and Definitions + Homonym using a weighting scheme: 
\begin{equation}
  \rho(u_1,u_2)=\begin{cases}
    \cos(u_1,u_2), & \text{if $u_1,u_2\:Related$}.\\
    0, & \text{otherwise}.
  \end{cases}
\end{equation}

For the evaluation, we calculate the Spearman correlation between human judgments and the labels computed as described. The results are presented in Table \ref{tab:lsc_results}, where we compare our model with XL-LEXEME, which represents the state of the art in this task, and other strategies for calculating the distance between definitions proposed by \citet{giulianelli-etal-2023-interpretable}. As the results show, the classification model achieves significantly better results compared to using metrics like BLEU or cosine distance applied to definition vectors calculated using SBERT. On the other hand, the model performs well below the state of the art, with a 15-point difference from XL-LEXEME. However, when we combine the classification model with XL-LEXEME, we achieve a new state-of-the-art result, indicating that the classification model produces reliable labels when classifying homonymy, i.e., instances annotated with 1.

\begin{table}[]
\resizebox{\columnwidth}{!}{%
\begin{tabular}{lc}
\textbf{Model} & \textbf{Correlation} \\
\hline
Definitions + SacreBLEU  & 0.108 \\
Definitions + METEOR  & 0.117 \\
Definitions +  Cosine similarity  & 0.264 \\
Definitions + Homonym  & 0.472 \\
XL-LEXEME & 0.623 \\
Definitions + Homonym  + XL-LEXEME & \textbf{0.646}
\end{tabular}%
}
\caption{Spearman correlation of human judgments vs model predictions, Definitions generated using the method of \citet{giulianelli-etal-2023-interpretable}}
\label{tab:lsc_results}
\end{table}

\subsection{Binary Change Task}
In SemEval-2020 Task 1, and in general for DWUGs datasets, words are annotated as stable or changed based on the removal and/or addition of a meaning (i.e., cluster). In this process, annotators' judgments are binarized, so that pairs of usages annotated with a score lower than 2.5 represent negative edges, while the rest of the usages represent positive edges. The optimization algorithm then seeks to group positive edges into clusters, keeping them separate from negative ones, with the resulting clusters representing word senses. 

Given the high correlation of the homonymy class with negative judgments (1 and 2), this represents an ideal proxy for automatically classifying words that have changed in meaning. Specifically, to classify words that have changed meaning in the SemEval dataset, we count for each word the occurrences of each label returned by the classification model, specifically the counts of hyperonym, hyponym, co-hyponym, antonym, and homonymy. We assign the label 1 (changed) to those words for which the homonymy class is the most frequent.
\begin{table}[]
\centering
\resizebox{0.8\columnwidth}{!}{%
\begin{tabular}{lc}
\textbf{Model} & \textbf{Accuracy} \\
\hline
Definitions + Homonym  & \textbf{0.783} \\
XL-LEXEME + 0.5 threshold & 0.761 \\
\hline
XL-LEXEME + Opt. threshold & 0.848 \\
\end{tabular}%
}
\caption{Binary task SemEval-2020 Task 1 (EN)}\label{tab:lsc_accuracy_results}
\end{table}

Table \ref{tab:lsc_accuracy_results} shows the accuracy on SemEval-2020 Task 1 - Subtask 1 for the English language. The results show that the Definitions + Homonym classification model achieved 78.3\% accuracy without the need for threshold adjustment. The XL-LEXEME model reached 76.1\% accuracy using the threshold of 0.5 (with which it has been trained), while optimizing the threshold improved its accuracy significantly to 84.8\%. It is notable that the Definitions + Homonym classification model performs well without the need to set up a threshold, which indicate that it is robust and less sensitive to parameter tuning compared to the XL-LEXEME model.

\section{Conclusion}
With this paper, we present a first computational approach to detecting the type of semantic change that a word has experienced. We release the LSC-CTD Benchmark, a first dataset for computational modeling of semantic change that is annotated by type of change. The benchmark is based on a digitized version of Blank's taxonomy extended with concise sense definitions. Because the resource is limited in size, it can only be used for testing purposes. To train a model for classifying change types, we used WordNet with synchronic sense definitions and relations between the synsets. 

The use of WordNet as a foundational resource for training our classifier enabled us to leverage a vast repository of semantic relationships and meanings. Our evaluation is grounded both in linguistic theory and computational metrics. We prove that homonyms are perceived by human annotators as the most distant semantic difference class in Blank's continuum of semantic proximity. 

We applied our classifier to extend semantic change detection models with type information and tested it on a standard evaluation benchmark, namely SemEval-2020 Task 1. We aimed to provide a more refined and accurate model for semantic change detection.  We found that by incorporating change types, in particular the notion of homonymy, we improved the state-of-the-art results for LSC on the graded task. For the binary task, we also saw encouraging results. 

Our findings underscore the importance of distinguishing between different types of semantic changes, such as generalization, specialization, and co-hyponymous transfer, in understanding the dynamics of semantic change.  In future work, we will extend our resources with more examples of change types, as well as additional synchronic resources.

\section{Limitations}
\vspace{-0.1cm}
Despite the promising outcomes, we acknowledge limitations in our study, regarding the employed resource, the limits of the distributional hypothesis and our focus on lexical type instead of token. 

First, Blank's database is a rich lexical resource, unique in its sort, but is nevertheless the work of one scholar working from one theoretical background. This likely  introduces bias in the selection and diversity of the examples. Another technical limitation of the benchmark is that we we took the generated definitions for granted as long as they were factually correct, but more detailed control on the specific wordings (for instance: to avoid specific lexical cues that bias the model in one or the other direction) might improve the results in the future.

Second, from a theoretical point of view, the types of semantic change under scrutiny do not emerge from text, but are the result of (oral) communicative exchanges and cognitive processes, and how both of these change over time. In other words, written input/context will always fall short of providing semantic information that is sufficiently rich to obtain optimal sense representations. Our choice of working with definitions partially circumvents this problem of under-determinism, but in general the limitation pertains to all kinds of context-based methods relying on the distributional hypothesis. As such, also our generated definitions are probably affected by it. 


Third, our initial approach to tackling the problem starts from lexical \textit{types} and their definition, for reasons of control and feasibility, but, as discussed above, semantic change unfolds in historical contexts and were likely, at the time of creation, selected from concrete communicative situations. The most appropriate level for modeling these phenomena is therefore at the level of lexical \textit{tokens}. 
To test this in the wild, we need to have manually annotated, historical resources in which change types are marked. This will enable studies to go from the level of definition to the level of usages. Our team is working on constructing such an annotated resource, but this is both expensive and time consuming and will likely only be at such a level that it can be used for testing purposes. 


\section*{Acknowledgments}
This work has in part been funded by the project Towards Computational Lexical Semantic Change Detection supported by the Swedish Research Council (2019–2022; contract 2018-01184), and in part by the research program Change is Key! supported by Riksbankens Jubileumsfond (under reference number M21-0021) and by an FWO junior postdoctoral grant (1281222N). The computational resources were provided by the National Academic Infrastructure for Supercomputing in Sweden (NAISS), partially funded by the Swedish Research Council through grant agreement no. 2022-06725.

We would also like to thank Haim Dubossarsky for providing valuable feedback on the preliminary draft of this work, as well as for engaging in early discussions that contributed to the development of this research.

\bibliography{anthology,custom}

\begin{thebibliography}{35}
\expandafter\ifx\csname natexlab\endcsname\relax\def\natexlab#1{#1}\fi

\bibitem[{Arefyev et~al.(2021)Arefyev, Homskiy, Fedoseev, Davletov, Protasov, and Panchenko}]{arefyev_deepmistake_2021}
Nikolay Arefyev, Daniil Homskiy, Maksim Fedoseev, Adis Davletov, Vitaly Protasov, and Alexander Panchenko. 2021.
\newblock {DeepMistake: Which Senses are Hard to Distinguish for a WordinContext Model}.
\newblock In \emph{Computational Linguistics and Intellectual Technologies - Papers from the Annual International Conference "Dialogue" 2021}, volume 2021-June.
\newblock Section: 20.

\bibitem[{Bevilacqua et~al.(2020)Bevilacqua, Maru, and Navigli}]{bevilacqua-etal-2020-generationary}
Michele Bevilacqua, Marco Maru, and Roberto Navigli. 2020.
\newblock \href {https://doi.org/10.18653/v1/2020.emnlp-main.585} {Generationary or {``}how we went beyond word sense inventories and learned to gloss{''}}.
\newblock In \emph{Proceedings of the 2020 Conference on Empirical Methods in Natural Language Processing (EMNLP)}, pages 7207--7221, Online. Association for Computational Linguistics.

\bibitem[{Blank(1997)}]{blank2012prinzipien}
Andreas Blank. 1997.
\newblock \emph{Prinzipien des lexikalischen Bedeutungswandels am Beispiel der romanischen Sprachen}, volume 285 of \emph{Beihefte zur Zeitschrift für romanische Philologie}.
\newblock Niemeyer, Tübingen.

\bibitem[{Bloomfield(1933)}]{bloomfield1994language}
Leonard Bloomfield. 1933.
\newblock \emph{Language}.
\newblock Holt, Rinehart, and Winston, New York.

\bibitem[{Br{\'e}al(1904)}]{breal1904essai}
Michel Br{\'e}al. 1904.
\newblock \emph{Essai de s{\'e}mantique (science des significations).}
\newblock Hachette.

\bibitem[{Cassotti et~al.(2023)Cassotti, Siciliani, DeGemmis, Semeraro, and Basile}]{cassotti-etal-2023-xl}
Pierluigi Cassotti, Lucia Siciliani, Marco DeGemmis, Giovanni Semeraro, and Pierpaolo Basile. 2023.
\newblock \href {https://doi.org/10.18653/v1/2023.acl-short.135} {{XL}-{LEXEME}: {W}i{C} pretrained model for cross-lingual {LEX}ical s{EM}antic chang{E}}.
\newblock In \emph{Proceedings of the 61st Annual Meeting of the Association for Computational Linguistics (Volume 2: Short Papers)}, pages 1577--1585, Toronto, Canada. Association for Computational Linguistics.

\bibitem[{Chung et~al.(2022)Chung, Hou, Longpre, Zoph, Tay, Fedus, Li, Wang, Dehghani, Brahma, Webson, Gu, Dai, Suzgun, Chen, Chowdhery, Castro-Ros, Pellat, Robinson, Valter, Narang, Mishra, Yu, Zhao, Huang, Dai, Yu, Petrov, Chi, Dean, Devlin, Roberts, Zhou, Le, and Wei}]{chung2022scaling}
Hyung~Won Chung, Le~Hou, Shayne Longpre, Barret Zoph, Yi~Tay, William Fedus, Yunxuan Li, Xuezhi Wang, Mostafa Dehghani, Siddhartha Brahma, Albert Webson, Shixiang~Shane Gu, Zhuyun Dai, Mirac Suzgun, Xinyun Chen, Aakanksha Chowdhery, Alex Castro-Ros, Marie Pellat, Kevin Robinson, Dasha Valter, Sharan Narang, Gaurav Mishra, Adams Yu, Vincent Zhao, Yanping Huang, Andrew Dai, Hongkun Yu, Slav Petrov, Ed~H. Chi, Jeff Dean, Jacob Devlin, Adam Roberts, Denny Zhou, Quoc~V. Le, and Jason Wei. 2022.
\newblock \href {http://arxiv.org/abs/2210.11416} {Scaling instruction-finetuned language models}.

\bibitem[{Darmesteter(1893)}]{darmesteter1893vie}
Ars{\`e}ne Darmesteter. 1893.
\newblock \emph{La vie des mots {\'e}tudi{\'e}e dans leurs significations}.
\newblock C. Delagrave.

\bibitem[{Fellbaum(1998)}]{princeton-wordnet}
Christiane Fellbaum. 1998.
\newblock \href {https://mitpress.mit.edu/9780262561167/} {\emph{WordNet: An Electronic Lexical Database}}.
\newblock MIT Press, Cambridge, MA.

\bibitem[{Gadetsky et~al.(2018)Gadetsky, Yakubovskiy, and Vetrov}]{gadetsky-etal-2018-conditional}
Artyom Gadetsky, Ilya Yakubovskiy, and Dmitry Vetrov. 2018.
\newblock \href {https://doi.org/10.18653/v1/P18-2043} {Conditional generators of words definitions}.
\newblock In \emph{Proceedings of the 56th Annual Meeting of the Association for Computational Linguistics (Volume 2: Short Papers)}, pages 266--271, Melbourne, Australia. Association for Computational Linguistics.

\bibitem[{Gardner et~al.(2022)Gardner, Khan, and Hung}]{defmodelingsurvey}
Noah Gardner, Hafiz Khan, and Chih-Cheng Hung. 2022.
\newblock \href {https://doi.org/10.3934/aci.2022005} {Definition modeling: literature review and dataset analysis}.
\newblock \emph{Applied Computing and Intelligence}, 2(1):83--98.

\bibitem[{Geeraerts(2010)}]{geeraerts2010}
Dirk Geeraerts. 2010.
\newblock \emph{Theories of Lexical Semantics}.
\newblock Oxford University Press, Oxford.

\bibitem[{Geeraerts(2020)}]{Geeraerts_2020}
Dirk Geeraerts. 2020.
\newblock \href {https://doi.org/https://doi.org/10.1002/9781118788516.sem042} {Semantic change. "what the smurf?"}.
\newblock In Daniel Gutzmann, Lisa Matthewson, Cécile Meier, Hotze Rullmann, and Thomas~E. Zimmermann, editors, \emph{The Wiley Blackwell Companion to Semantics}. Wiley Blackwell, Hoboken NJ.

\bibitem[{Giulianelli et~al.(2023)Giulianelli, Luden, Fernandez, and Kutuzov}]{giulianelli-etal-2023-interpretable}
Mario Giulianelli, Iris Luden, Raquel Fernandez, and Andrey Kutuzov. 2023.
\newblock \href {https://doi.org/10.18653/v1/2023.acl-long.176} {Interpretable word sense representations via definition generation: The case of semantic change analysis}.
\newblock In \emph{Proceedings of the 61st Annual Meeting of the Association for Computational Linguistics (Volume 1: Long Papers)}, pages 3130--3148, Toronto, Canada. Association for Computational Linguistics.

\bibitem[{Hamilton et~al.(2016)Hamilton, Leskovec, and Jurafsky}]{hamilton2016cultural}
William~L Hamilton, Jure Leskovec, and Dan Jurafsky. 2016.
\newblock Cultural shift or linguistic drift? comparing two computational measures of semantic change.
\newblock In \emph{Proceedings of the Conference on Empirical Methods in Natural Language Processing. Conference on Empirical Methods in Natural Language Processing}, volume 2016, page 2116. ACL.

\bibitem[{Huang et~al.(2021)Huang, Kajiwara, and Arase}]{huang-etal-2021-definition}
Han Huang, Tomoyuki Kajiwara, and Yuki Arase. 2021.
\newblock \href {https://doi.org/10.18653/v1/2021.emnlp-main.194} {Definition modelling for appropriate specificity}.
\newblock In \emph{Proceedings of the 2021 Conference on Empirical Methods in Natural Language Processing}, pages 2499--2509, Online and Punta Cana, Dominican Republic. Association for Computational Linguistics.

\bibitem[{Karaman(2008)}]{karaman2008}
Burcu~I. Karaman. 2008.
\newblock \href {https://doi.org/10.1093/ijl/ecn011} {{On Contronymy1}}.
\newblock \emph{International Journal of Lexicography}, 21(2):173--192.

\bibitem[{Laicher et~al.(2021)Laicher, Kurtyigit, Schlechtweg, Kuhn, and im~Walde}]{laicher}
Severin Laicher, Sinan Kurtyigit, Dominik Schlechtweg, Jonas Kuhn, and Sabine~Schulte im~Walde. 2021.
\newblock \href {https://doi.org/10.18653/v1/2021.eacl-srw.25} {Explaining and improving {BERT} performance on lexical semantic change detection}.
\newblock In \emph{Proc. of the 16th Conference of the European Chapter of the Association for Computational Linguistics: Student Research Workshop, {EACL} 2021, Online, April 19-23, 2021}, pages 192--202. Association for Computational Linguistics.

\bibitem[{Liu et~al.()Liu, Ott, Goyal, Du, Joshi, Chen, Levy, Lewis, Zettlemoyer, and Stoyanov}]{liuroberta2019}
Yinhan Liu, Myle Ott, Naman Goyal, Jingfei Du, Mandar Joshi, Danqi Chen, Omer Levy, Mike Lewis, Luke Zettlemoyer, and Veselin Stoyanov.
\newblock {RoBERTa: A Robustly Optimized BERT Pretraining Approach}.
\newblock \emph{{arxiv}}, abs/1907.11692.

\bibitem[{Maudslay and Teufel(2022)}]{maudslay-teufel-2022-metaphorical}
Rowan~Hall Maudslay and Simone Teufel. 2022.
\newblock \href {https://aclanthology.org/2022.coling-1.7} {Metaphorical polysemy detection: Conventional metaphor meets word sense disambiguation}.
\newblock In \emph{Proceedings of the 29th International Conference on Computational Linguistics}, pages 65--77, Gyeongju, Republic of Korea. International Committee on Computational Linguistics.

\bibitem[{Mickus et~al.(2022)Mickus, van Deemter, Constant, and Paperno}]{DBLP:conf/semeval/MickusDCP22}
Timothee Mickus, Kees van Deemter, Mathieu Constant, and Denis Paperno. 2022.
\newblock \href {https://doi.org/10.18653/V1/2022.SEMEVAL-1.1} {Semeval-2022 task 1: {CODWOE} - comparing dictionaries and word embeddings}.
\newblock In \emph{Proceedings of the 16th International Workshop on Semantic Evaluation, SemEval@NAACL 2022, Seattle, Washington, United States, July 14-15, 2022}, pages 1--14. Association for Computational Linguistics.

\bibitem[{Miller(2013)}]{miller2013}
Justin~J Miller. 2013.
\newblock Graph database applications and concepts with neo4j.
\newblock In \emph{Proceedings of the southern association for information systems conference, Atlanta, GA, USA}, volume 2324.

\bibitem[{Montanelli and Periti(2023)}]{montanelli2023survey}
Stefano Montanelli and Francesco Periti. 2023.
\newblock \href {https://doi.org/https://doi.org/10.48550/arXiv.2304.01666} {{A Survey on Contextualised Semantic Shift Detection}}.

\bibitem[{Navigli(2009)}]{navigli_survey}
Roberto Navigli. 2009.
\newblock \href {https://doi.org/10.1145/1459352.1459355} {{Word Sense Disambiguation: A Survey}}.
\newblock \emph{ACM Comput. Surv.}, 41(2).

\bibitem[{Paul(1880)}]{paul2010prinzipien}
Hermann Paul. 1880.
\newblock \emph{Prinzipien der Sprachgeschichte}.
\newblock Niemeyer, Halle.

\bibitem[{Periti and Tahmasebi(2024)}]{periti2024chatgpt}
Francesco Periti and Nina Tahmasebi. 2024.
\newblock \href {https://doi.org/10.48550/arXiv.2402.12011} {{A Systematic Comparison of Contextualized Word Embeddings for Lexical Semantic Change}}.
\newblock In \emph{Proceedings of the 2024 Conference of the North American Chapter of the Association for Computational Linguistics: Human Language Technologies}, Mexico City, Mexico. Association for Computational Linguistics.

\bibitem[{Rachinskiy and Arefyev(2021)}]{rachinskiy_zeroshot_2021}
Maxim Rachinskiy and Nikolay Arefyev. 2021.
\newblock {Zeroshot Crosslingual Transfer of a Gloss Language Model for Semantic Change Detection}.
\newblock In \emph{Computational Linguistics and Intellectual Technologies - Papers from the Annual International Conference "Dialogue" 2021}, volume 2021-June.
\newblock Section: 20.

\bibitem[{Reisig(1839)}]{reisig1839professor}
Karl~Christian Reisig. 1839.
\newblock \emph{Professor K. Reisig's Vorlesungen {\"u}ber lateinische Sprachwissenschaft}.
\newblock Verlag der Lehnhold'schen Buchhandlung.

\bibitem[{Schlechtweg et~al.(2018)Schlechtweg, im~Walde, and Eckmann}]{schlechtweg2018diachronic}
Dominik Schlechtweg, Sabine~Schulte im~Walde, and Stefanie Eckmann. 2018.
\newblock \href {https://doi.org/10.18653/v1/n18-2027} {{Diachronic Usage Relatedness (DURel): {A} Framework for the Annotation of Lexical Semantic Change}}.
\newblock In \emph{Proceedings of the 2018 Conference of the North American Chapter of the Association for Computational Linguistics: Human Language Technologies, NAACL-HLT, Volume 2 (Short Papers)}, pages 169--174, New Orleans, Louisiana, USA. Association for Computational Linguistics.

\bibitem[{Schlechtweg et~al.(2020)Schlechtweg, McGillivray, Hengchen, Dubossarsky, and Tahmasebi}]{schlechtweg2020semeval}
Dominik Schlechtweg, Barbara McGillivray, Simon Hengchen, Haim Dubossarsky, and Nina Tahmasebi. 2020.
\newblock \href {https://www.aclweb.org/anthology/2020.semeval-1.1/} {{SemEval-2020 Task 1: Unsupervised Lexical Semantic Change Detection}}.
\newblock In \emph{Proceedings of the Fourteenth Workshop on Semantic Evaluation, SemEval@COLING2020}, pages 1--23, Barcelona (online). International Committee for Computational Linguistics.

\bibitem[{Schlechtweg et~al.(2021)Schlechtweg, Tahmasebi, Hengchen, Dubossarsky, and McGillivray}]{schlechtweg-resource-2020-21}
Dominik Schlechtweg, Nina Tahmasebi, Simon Hengchen, Haim Dubossarsky, and Barbara McGillivray. 2021.
\newblock {DWUG: A} large {R}esource of {D}iachronic {W}ord {U}sage {G}raphs in {F}our {L}anguages.
\newblock In \emph{{Annual Conference of the North American Chapter of the Association for Computational Linguistics, (NAACL-HLT 2021)}}, Mexico City, Mexico. Association for Computational Linguistics.

\bibitem[{Stern(1931)}]{stern1931meaning}
Gustaf Stern. 1931.
\newblock \emph{Meaning and change of meaning; with special reference to the English language.}
\newblock Indiana University Press, Bloomington.

\bibitem[{Tahmasebi et~al.(2021)Tahmasebi, Borin, and Jatowt}]{tahmasebi2021survey}
Nina Tahmasebi, Lars Borin, and Adam Jatowt. 2021.
\newblock \href {https://doi.org/10.5281/zenodo.5040302} {\emph{{Survey of Computational Approaches to Lexical Semantic Change Detection}}}, pages 1--91. Language Science Press, Berlin.

\bibitem[{Traugott(2017)}]{traugott2017}
Elizabeth~Closs Traugott. 2017.
\newblock \href {https://doi.org/10.1093/acrefore/9780199384655.013.323} {Semantic change}.
\newblock In \emph{Oxford Research Encyclopedia of Linguistics}. Oxford University Press.

\bibitem[{Ullmann(1957)}]{ullmann}
S.~Ullmann. 1957.
\newblock \href {https://books.google.it/books?id=YOZYAAAAMAAJ} {\emph{The Principles of Semantics}}.
\newblock Glasgow University publications. Jackson.

\end{thebibliography}
\balance
\clearpage
\appendix

\begin{table}[]
\centering
\resizebox{\columnwidth}{!}{%
\begin{tabular}{lrrr}
\toprule
 & \multicolumn{1}{c}{Train} & \multicolumn{1}{c}{Development} & \multicolumn{1}{c}{Test} \\ \cline{2-4} 
Unique synsets & 79469 & 15994 & 15978 \\
Definition pairs & 123034 & 12379 & 12380 \\
Hyponyms pairs & 30000 & 3000 & 3000 \\
Hyperonyms pairs & 30000 & 3000 & 3000 \\
Co-hyponyms pairs & 30000 & 3000 & 3000 \\
Homonyms pairs & 30000 & 3000 & 3000 \\
Antonyms pairs & 3033 & 379 & 380 \\
Noun pairs & 85084 & 8535 & 8480 \\
Verb pairs & 20772 & 2032 & 2137 \\
Adjective pairs & 14268 & 1491 & 1473 \\
Adverb pairs & 2910 & 321 & 290\\
\bottomrule
\end{tabular}%
}
\captionof{table}{WN statistics.}
\label{tab:wn_stats}
\end{table}

\section{LSC-CTD Benchmark}
\label{app:LSC-def-bench}

In Table \ref{tab:lsc_type_benchmark}, we present an excerpt from the LSC-CTD Benchmark, a manually curated dataset of dictionary-like definition pairs for words that experienced semantic change. It includes Blank's original annotations detailing the target word followed by a language tag (lt: Latin, vlt: Vulgar Latin) and German glosses of the original, primary meaning- (Old Meaning) and of the derived, secondary meaning (New Meaning). We translate Blank's glosses into English using Google Translate's API (between double quotation marks). Following this, we use the prompt shown in Figure \ref{fig:gpt-prompt} to generate definitions that are more akin to those found in a dictionary, employing \texttt{gpt-4} API (between single quotation marks). This prompt includes some examples we initially set up, with the actual generation occurring through few-shot learning. Furthermore Blank's dataset contains the alleged cause of the semantic change and, not shown here, the higher-level association relation that exists at the conceptual level. Finally, a linguist with expertise in historical linguistics and semantics manually reviewed and refined the collection of generated translations and definitions, replacing or modifying them as necessary.

\begin{table*}[htb!]
\centering
\resizebox{0.95\textwidth}{!}{%
\begin{tabular}{|lp{5cm}p{5cm}ll|}
\hline
\textbf{Word} & \textbf{Old Meaning} & \textbf{New meaning} & \textbf{Cause} & \textbf{Type} \\
\hline
\textit{*adripare:vlt} & am Ufer ankommen & ankommen & prototype / frame & generalization \\
 &  "arrive at the bank/shore" &  "arrive" &  &  \\
 &  \lq{arrive at the bank of a river or the shore of a lake or sea}\rq & \lq{to reach a place, especially at the end of a journey}\rq &  &  \\
\hline
\textit{necare:lt} & töten & ertränken & socio-cultural change & specialization \\
 & "kill" & "drown" &  &  \\
 & \lq{to cause the death of a living thing, typically involving an act of violence or an intention to harm.}\rq & \lq{to cause to die by submersion in liquid, especially by forcing the head under the water.}\rq &  &  \\
\hline
\textit{*ratta} & Ratte & Maus & referential vagueness & co-hyponymous transfer \\
 & "rat" & "mouse" &  &  \\
 & \lq{a small rodent, larger than a mouse, that has a long tail and is considered to be harmful}\rq & \lq{a small mammal with short fur, a pointed face, and a long tail}\rq  &  &  \\
\hline
\textit{sacer:lt} & heilig , geheiligt & verflucht & taboo & auto-antonymy \\
 & "sacred" & "cursed" &  &  \\
 & \lq{considered to be holy and deserving respect, especially because of a connection with a god}\rq & \lq{experiencing bad luck caused by a magic curse}\rq &  & \\
\hline
\end{tabular}%
}
\captionof{table}{Snippet of the LSC Cause-Type-Definitions Benchmark}
\label{tab:lsc_type_benchmark}
\end{table*}

\section{Baseline}
\label{app:baseline}
As our baselines, we consider two models. The first is naturally the base version of roberta (\texttt{roberta-base}) \cite{liuroberta2019}, as we have utilized the large version for the experiments reported in the paper (see Figure 5). The training of the base model is identical to that of the large model, employing the same hyperparameters and data. The second baseline, on the other hand, is based on the tf-idf weighting model (see Figure 4). Specifically, we calculate weights across all definitions present in the training set. Subsequently, we extract vectors for each definition and use the concatenation of vectors from pairs of definitions as features for our model. In particular, we employ a simple classifier based on stochastic gradient descent (SGD), utilizing the implementation provided by scikit-learn\footnote{\url{https://scikit-learn.org/stable/modules/generated/sklearn.linear_model.SGDClassifier.html}} with default parameters, except for class weighting where we attempt to mitigate imbalance using the balanced parameter. The tf-idf baseline is used to investigate trivial patterns among pairs of definitions, that is, whether the simple co-occurrence of words can be a signal for identifying the target class.
Not many differences are observed between \texttt{roberta-base} and \texttt{roberta-large}, both on the WN test set and the LSC-CTD benchmark. In \texttt{roberta-large} there is however a relatively large improvement in the detection of auto-antonymy (+ 0.12) and generalization (+ 0.03), but also a lower performance for co-hyponymous transfer (-0.04). The comparison between the \texttt{roberta} models and the tf-idf models is more striking, with the latter model failing on differentiating homonymy from auto-antonymy (probably due to not taking into account the syntactic position of negations) again on both test sets. Surprisingly, regarding the identification of co-hyponymous transfer, the tf-idf model performs much better (+0.16 relative to \texttt{roberta-large}).

\section{WN Statistics}
\label{app:wn_stats}
In Table \ref{tab:wn_stats} we report the statistics we use to train the sense relationship model. The table includes for each split (train, dev, and test set) the unique number of synsets, the number of pairs of definitions, and the number of pairs of definitions for each class (hyponym, hyperonym, co-hyponym, homonym, and antonym). Additionally, the number of distinct pairs for each part of speech (nouns, verbs, adjectives, and adverbs) is reported.

\begin{figure*}[h]
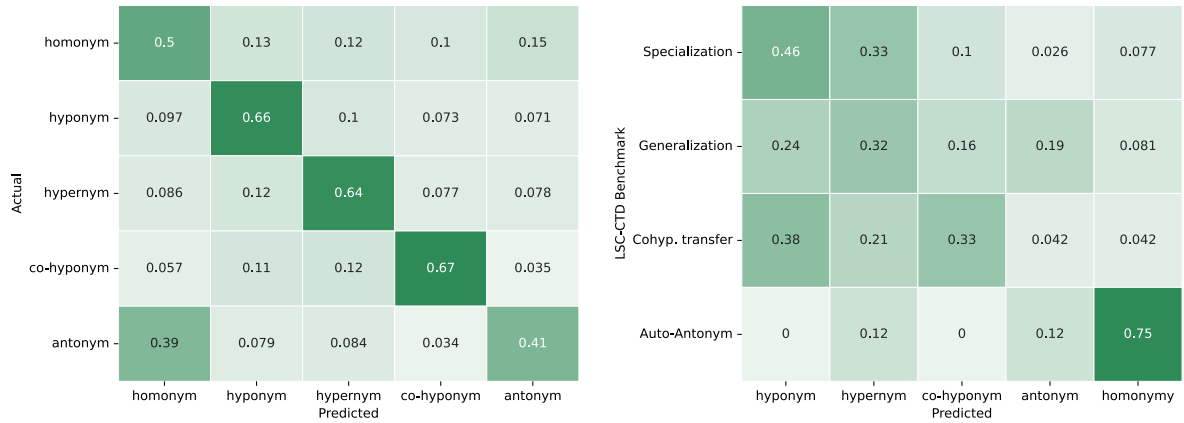

\begin{subfigure}[b]{0.49\textwidth}
    \centering
    \includesvg[width=\columnwidth]{svm_cfn_wn.svg}
    \caption{WN test set.}
    \label{fig:svm_wn}
\end{subfigure}
\begin{subfigure}[b]{0.49\textwidth}
    \centering
    \includesvg[width=\columnwidth]{svm_cfn.svg}
    \caption{LSC-CTD Benchmark.}
    \label{fig:svm_blank}
\end{subfigure}
\caption{SGD TF-IDF.}
\label{ref:svm}
\end{figure*}

\begin{figure*}[h]
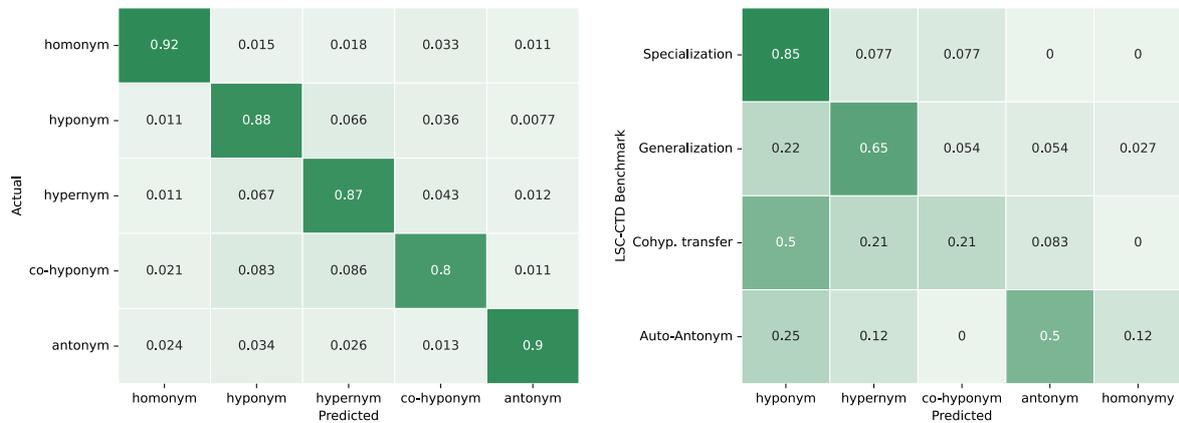

\begin{subfigure}[b]{0.49\textwidth}
    \centering
    \includesvg[width=\columnwidth]{base_cfn_wn.svg}
    \caption{WN test set.}
    \label{fig:roberta_small_wn}
\end{subfigure}
\begin{subfigure}[b]{0.49\textwidth}
    \centering
    \includesvg[width=\columnwidth]{base_cfn.svg}
    \caption{LSC-CTD Benchmark.}
    \label{fig:roberta_small_blank}
\end{subfigure}
\caption{roberta-base.}
\label{ref:roberta_small}
\end{figure*}

\begin{figure*}[htbp]
    \centering
    \begin{subfigure}[t]{\textwidth}
        \includesvg[width=\columnwidth]{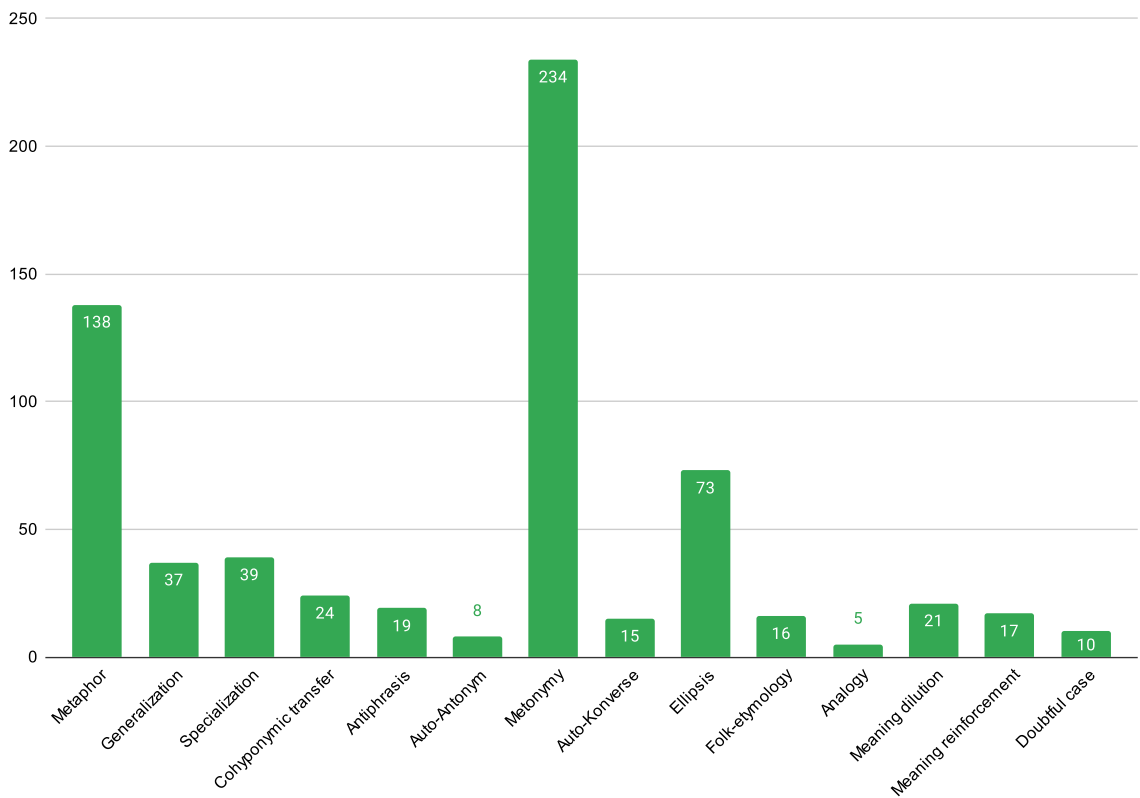}
        \caption{Number of cases for each type.}
        \label{fig:type_stats}
    \end{subfigure}
    \begin{subfigure}[t]{\textwidth}
        \includesvg[width=\columnwidth]{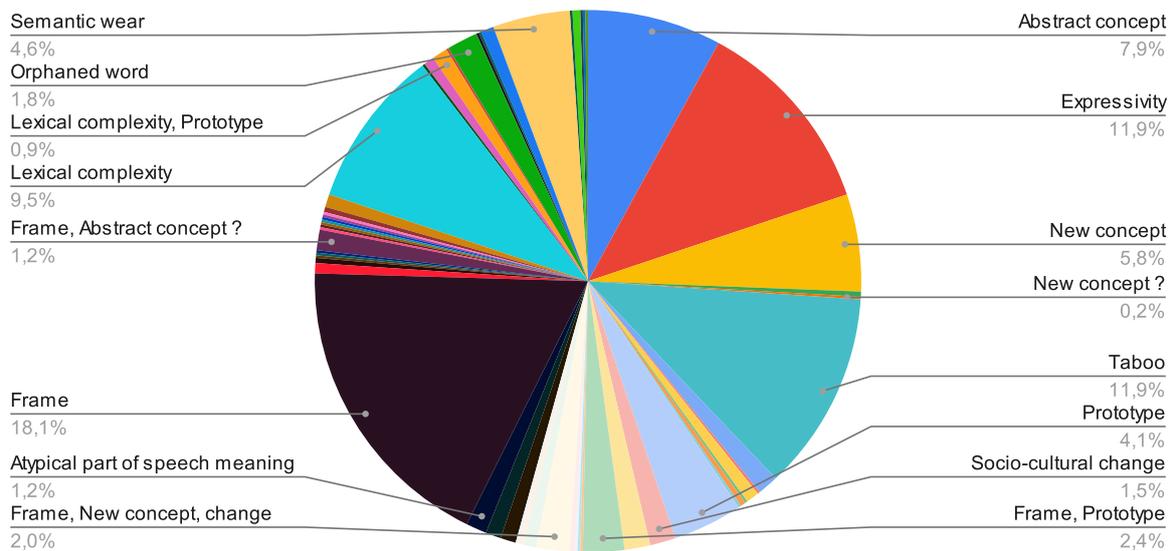}
        \caption{Number of cases for each cause.}
        \label{fig:cause_stats}
    \end{subfigure}
    \caption{LSC CTD Benchmark statistics.}
    \label{fig:lsc-ctd_stats}
\end{figure*}

\begin{figure*}
\centering
    \includesvg[width=\textwidth]{prompt.svg}
    \captionof{figure}{Chat GPT prompt.}
    \label{fig:gpt-prompt}
\end{figure*}

\end{document}